\pdfoutput=1
\documentclass[11pt]{article}

\usepackage{hyperref}

\makeatletter

\def\showhyphens#1{}%
\makeatother

\usepackage[final]{acl}

\usepackage{times}
\usepackage{latexsym}
\usepackage{graphicx}
\usepackage{placeins}
\usepackage{natbib}
\usepackage{stfloats} 
\usepackage{lipsum}
\usepackage{booktabs}
\usepackage{amssymb}
\usepackage[T1]{fontenc}
\usepackage[utf8]{inputenc}
\usepackage{microtype}

\usepackage{inconsolata}

\usepackage{amsmath}
\usepackage{accents}

\title{CMV-Fuse: Cross Modal-View Fusion of AMR, Syntax, and Knowledge Representations for Aspect Based Sentiment Analysis}
\author{Smitha Muthya Sudheendra\thanks{Equal contribution.}\\
  University of Minnesota, Twin Cities\\
  \texttt{muthy009@umn.edu} \\\And
  Mani Deep Cherukuri\footnotemark[1]  \\
  University of Minnesota, Twin Cities\\
  \texttt{cheru050@umn.edu} \\\AND
  Jaideep Srivastava \\
  University of Minnesota, Twin Cities\\
  \texttt{srivasta@umn.edu}}

\begin{document}
\maketitle

\begin{abstract}
Natural language understanding inherently depends on integrating multiple complementary perspectives spanning from surface syntax to deep semantics and world knowledge. However, current Aspect-Based Sentiment Analysis (ABSA) systems typically exploit isolated linguistic views, thereby overlooking the intricate interplay between structural representations that humans naturally leverage. We propose CMV-Fuse, a Cross-Modal View fusion framework that emulates human language processing by systematically combining multiple linguistic perspectives. Our approach systematically orchestrates four linguistic perspectives: Abstract Meaning Representations, constituency parsing, dependency syntax, and semantic attention, enhanced with external knowledge integration. Through hierarchical gated attention fusion across local syntactic, intermediate semantic, and global knowledge levels, CMV-Fuse captures both fine-grained structural patterns and broad contextual understanding. A novel structure aware multi-view contrastive learning mechanism ensures consistency across complementary representations while maintaining computational efficiency. Extensive experiments demonstrate substantial improvements over strong baselines on standard benchmarks, with analysis revealing how each linguistic view contributes to more robust sentiment analysis.

\end{abstract}
\section{Introduction}
Aspect-based sentiment analysis (ABSA) is a fine-grained task in sentiment analysis that aims to identify the sentiment polarity of specific aspects within a sentence. For example, in the sentence \textit{"The small dish was delicious,"} ABSA must determine that \textit{"dish"} has a mixed sentiment - negative from \textit{"small"} (referring to portion size) and positive from "delicious" (referring to taste). This task requires a deep understanding of aspect-opinion relationships, making it a critical component of natural language understanding research, with practical applications in customer feedback analysis, opinion mining, and other domains.

\begin{figure}[ht!]
    \centering
    \includegraphics[width=0.9\linewidth]{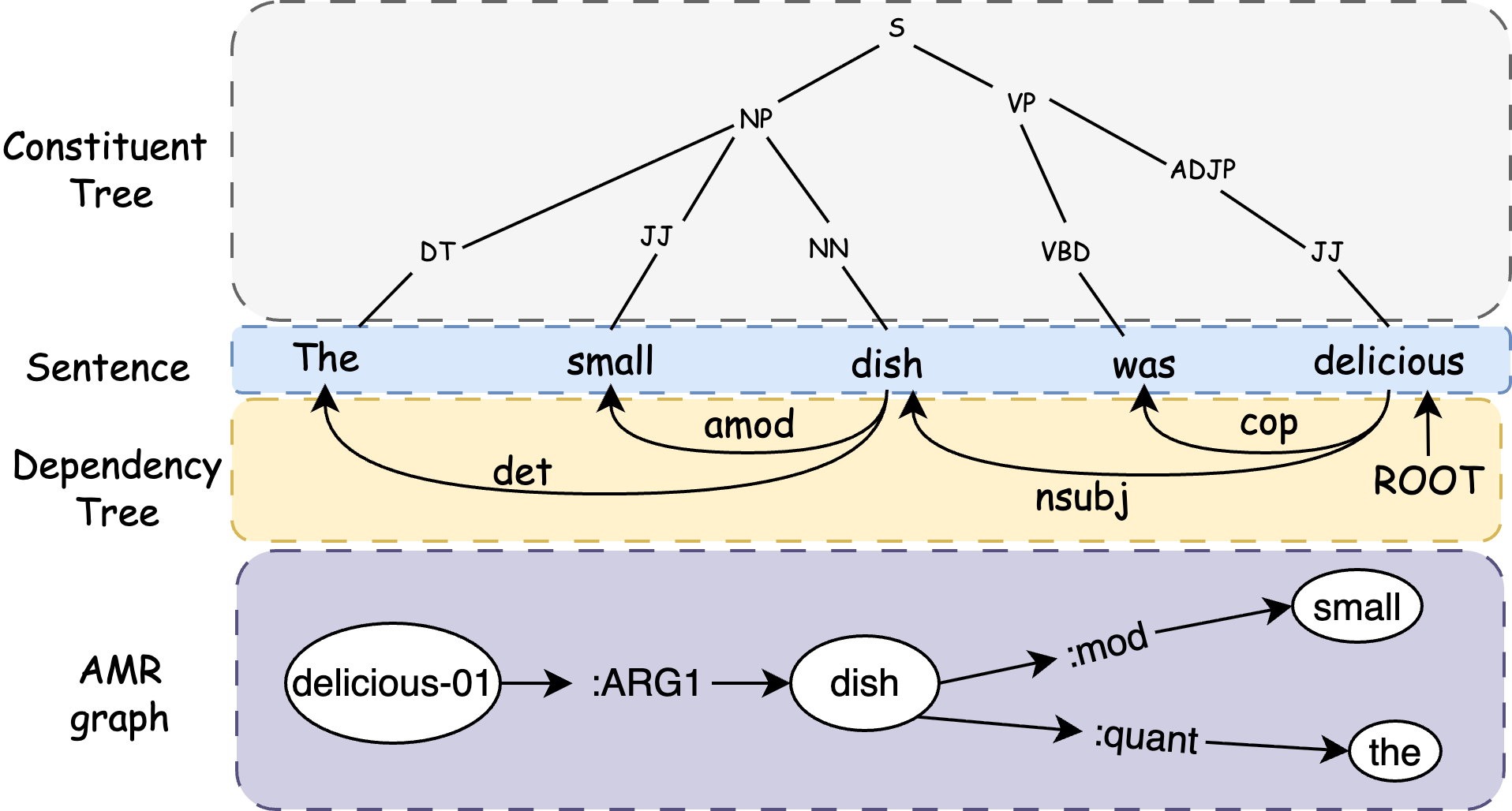}
    \caption{Constituent Tree, Dependency Tree and AMR graph}
    \label{fig:Introduction}
\end{figure}

Context-based methods struggle to associate multiple opinion terms with an aspect simultaneously (e.g., linking \textit{``small''} and \textit{``delicious''} to \textit{``dish''}), while syntactic structure-based approaches suffer from a critical \textit{semantic--syntactic mismatch}. As shown in Figure~\ref{fig:Introduction}, surface-level syntactic structures often fail to capture sentiment-bearing semantic relations: although \textit{``small''} is attached to \textit{``dish''} via the \textit{amod} relation, its grammatical position can mislead models, as it resides within the subject phrase governed by \textit{``was''} rather than reflecting its semantic role. In contrast, the AMR graph explicitly encodes \textit{``delicious-01''} with \textit{dish} as \textit{ARG1} and represents modifiers such as \textit{``small''} semantically. However, existing knowledge-enhanced methods struggle to integrate such structured information efficiently, often relying on costly subgraph sampling or weak cross-view alignment, limiting their ability to distinguish sentiment attributes such as portion size and taste.

Despite recent progress, existing sentiment models lack a principled mechanism for integrating heterogeneous linguistic representations that encode complementary sentiment cues. Most prior work either optimizes a single linguistic view (e.g., syntax, semantics, or attention) or relies on ad hoc fusion strategies such as feature concatenation or task-specific heuristics, which limits cross-view alignment and generalization. Consequently, these models struggle to jointly reason over global semantic abstractions (e.g., AMR), hierarchical phrase structure (constituency), grammatical relations (dependency), and contextual associations (semantic attention), especially in complex aspect-level scenarios.
To address this gap, we propose CMV-Fuse\footnote{We will release the code upon acceptance}, a Cross-Modal View fusion framework that independently encodes complementary linguistic views and aligns them through a hierarchical fusion architecture. CMV-Fuse adopts a modular, plug-and-play design that enables state-of-the-art single-view encoders to be systematically integrated, while explicitly modeling cross-view interactions to strengthen aspect–opinion association.

Our main contributions are as follows:
\begin{enumerate}
    \item We introduce CMV-Fuse, a unified plug and play architecture that synergizes AMR semantics, constituency parsing, dependency syntax, and semantic attention through cross-view consistency constraints to bridge the semantic-syntactic gap.
    \item We propose a three-level hierarchical gated fusion mechanism that orchestrates complementary linguistic perspectives across local syntactic, intermediate semantic, and global knowledge levels within a unified architecture.
    \item We introduce a multi-view structure-aware contrastive learning objective to align the representations, enhancing cross-modal consistency without added complexity.
    \item Evaluation and ablation study on three ABSA benchmark datasets demonstrating that CMV-Fuse outperforms strong baselines, revealing how contrastive alignment of multiple linguistic views outperforms single structural perspectives.
\end{enumerate}

\section{Related Works}

\subsection{Deep Learning Approaches for ABSA}
Early context-based methods focused on LSTMs and attention mechanisms. \citet{tang-etal-2016-effective} introduced Target-Dependent LSTM (TD-LSTM), and \citet{wang-etal-2016-attention} pioneered attention-based LSTM (ATAE-LSTM).  \citet{ma2017interactiveattentionnetworksaspectlevel} advanced this with Interactive Attention Networks (IAN). However, these methods suffer from an aspect-opinion misalignment and limited structural awareness.

Graph-based syntactic methods emerged to capture structural relationships. \citet{zhang-etal-2019-aspect} pioneered AspectGCN by applying Graph Convolutional Networks to dependency trees with aspect-focused edge pruning. \citet{sun-etal-2019-aspect} extended this with Aspect-Specific GCN (ASGCN) using learnable edge weights for aspect-specific dependency adaptation. \citet{liang2022bisyn} proposed the combination of constituent and dependency trees in ABSA to capture syntactic dependencies in both bottom-up and top-down directions using attention mechanisms, enabling richer contextual representations for each token. \citet{li-etal-2021-dual-graph} introduced Dual-GCN with parallel syntactic and semantic graph encoders connected via BiAffine transformation. \citet{yang-etal-2020-constituency} first explored constituency parsing for ABSA, demonstrating complementary phrase-level information to dependency trees. Despite advances, syntactic methods face critical semantic-syntactic mismatch where grammatical dependencies fail to align with sentiment-bearing relationships. Semantic approaches, such as Abstract Meaning Representation (AMR), provide abstraction beyond syntax. \citet{ma-etal-2023-amr} introduced APARN, which improves aspect-opinion association through semantic roles. However, AMR-based approaches for ABSA remain underexplored.

\subsection{Multi-View Fusion and Knowledge Integration}
Recent work recognizes that single linguistic views provide incomplete information. \cite{ZHOU2020106292} pioneered knowledge graph sampling with ConceptNet subgraphs for CommonsenseGCN, while \citet{Zhong_2023}  developed Knowledge Graph Augmented Network (KGAN) addressing computational complexity through efficient attention-based knowledge fusion. However, these approaches face knowledge-text alignment issues where external representations often fail to align semantically with textual features.

Multi-view fusion approaches attempt systematic integration of complementary representations. \citet{li-etal-2021-dual-graph} combined syntactic and semantic graphs through BiAffine transformation, while recent work also explored contrastive learning for structural consistency. In ABSA, \citet{10.1145/3459637.3482096} applied contrastive learning to separate sentiment features based on polarity and patterns, and \citet{LI2023110648} developed aspect-aware contrastive learning to enforce consistency between different structural views. Despite progress, existing approaches remain limited to two or three views with simple fusion strategies, lacking scalable frameworks for systematic multi-view integration 

Although these prior ABSA models incorporate multiple linguistic features, their fusion mechanisms are often handcrafted or tightly coupled to specific representations, limiting extensibility and cross-view alignment. CMV-Fuse differs by introducing a modular and hierarchical fusion framework that systematically aligns heterogeneous linguistic views, allowing complementary information to be integrated without redesigning task-specific architectures.

\section{Model Overview}

In the ABSA task, the goal is to predict the sentiment polarity of an aspect term within a sentence. Let $s = \{w_1, w_2, \ldots, w_n\}$ represent a sentence with $n$ words, and $a = \{a_1, a_2, \ldots, a_m\}$ denote an aspect term, where $a$ is a subsequence of $s$. The objective is to determine the sentiment polarity $c_a \in \{Positive, Neutral, Negative\}$ for the aspect $a$.

\begin{figure*}[ht!]
    \centering
    \includegraphics[width=0.9\linewidth]{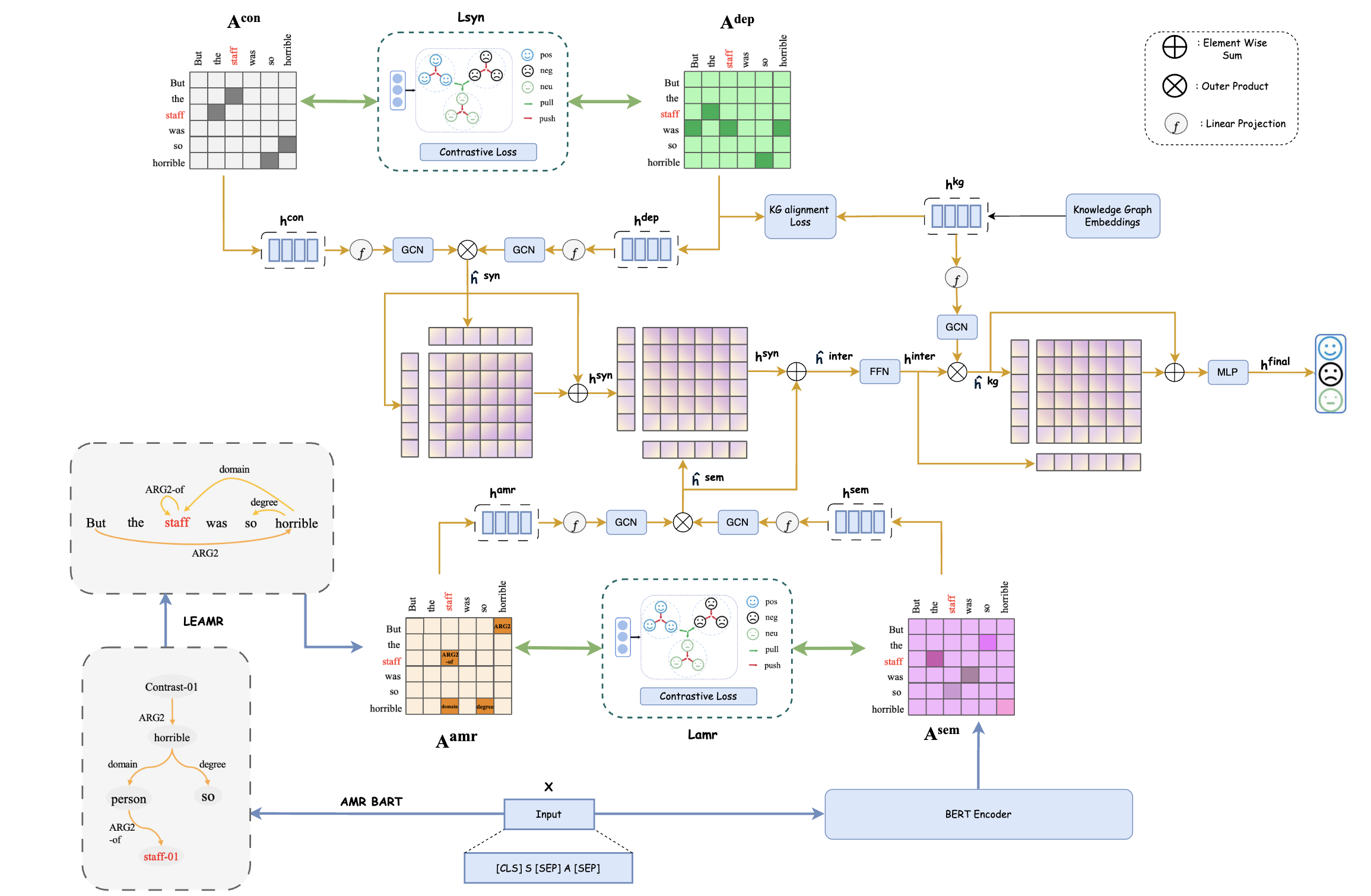}
    \caption{Model Architecture}
    \label{fig:model_framework}
\end{figure*}

As shown in Figure \ref{fig:model_framework}, our model consists of three key components: (1) Multi-View Graph Encoder, (2) Hierarchical Cross-Modal Fusion, and (3) Multi-view structure-aware Contrastive Alignment. The process begins with BERT contextualized representations $H^{\text{BERT}} = \{h^{\text{BERT}}_1, h^{\text{BERT}}_2, \ldots, h^{\text{BERT}}_n \} \in \mathbb{R}^{n \times d}$, obtained from sentence–aspect pairs formatted as $x = \text{[CLS]}\, s \,\text{[SEP]}\, a \,\text{[SEP]}$, which serve as initial node features for the graph encoders that process Abstract Meaning Representations, constituency parsing, dependency syntax, and semantic attention to capture structural and contextual patterns.

\subsection{Multi-View Graph Encoder Module}

\textbf{AMR Adjacency Matrix}: Abstract Meaning Representation (AMR) provides a semantic abstraction that captures the core meaning of sentences while abstracting away surface syntactic variations. Our CMV framework incorporates AMR graphs as a complementary representation for aspect-based sentiment analysis. We utilize AMR-BART \cite{bai-etal-2022-graph} to generate AMR graphs from input sentences, converting them into token-aligned adjacency matrices via concept-to-token mapping using LEAMR \cite{blodgett-schneider-2021-probabilistic}.

The token-level AMR adjacency matrix is constructed using an edge vocabulary $\mathcal{V}_{AMR}$, which includes semantic relations such as argument roles (:ARG0, :ARG1, :ARG2), modifiers (:mod, :manner, :time), operators (:op1, :op2), and prepositional relations (:prep-*). This vocabulary handles relation transformations, including inverse relations (e.g., `:ARG0-of` $\rightarrow$ `:ARG0`) and prepositional relations (`:prep-X` $\rightarrow$ `X`).

Let $\mathcal{E}_{\text{amr}} = \{(s, r, t) \mid s, t \in [0, n), r \in \mathcal{V}_{\text{amr}}\}$ represent the AMR edge set, where $s, t$ are token indices and $r$ is the relation label. The adjacency matrix $\mathbf{A}^{\text{amr}} \in \mathbb{Z}^{n \times n}$ is defined as:
\begin{equation}
\mathbf{A}^{\text{amr}}_{ij} = \begin{cases}
\mathcal{V}_{\text{amr}}(r) & \text{if edge } (i, r, j) \in \mathcal{E}_{\text{amr}} \\
\mathcal{V}_{\text{amr}}(\text{self}) & \text{if } i = j \\
\mathcal{V}_{\text{amr}}(\text{none}) & \text{otherwise}
\end{cases}
\end{equation}

Initially, the matrix is set to $\mathcal{V}_{\text{amr}}(\text{none})$ for all entries. Directed edges are populated with corresponding relation indices $\mathcal{V}_{\text{amr}}(r)$, and the diagonal entries are set to $\mathcal{V}_{\text{amr}}(\text{self})$ to form self-loops.

\textbf{Dependency Adjacency Matrix}: To exploit syntactic dependencies, we construct a dependency parse tree $\mathcal{T} = (\mathcal{V}_T, \mathcal{E}_T)$, where $\mathcal{V}_T$ represents the set of tokens, and $\mathcal{E}_T$ denotes the set of directed dependency edges. For each token $w_i$, its syntactic governor is defined as $\text{head}(w_i)$. The undirected dependency graph $\mathcal{G}_{dep} = (\mathcal{V}, \mathcal{E}_{dep})$ is derived by converting directed dependencies into symmetric edges while maintaining the syntactic relationships.

The adjacency matrix $\mathbf{A}^{dep} \in \{0, 1\}^{n \times n}$ for the dependency graph is defined as:

\begin{equation}
\mathbf{A}^{dep}_{ij} = \begin{cases}
1 & \text{if } (w_i, w_j) \in \mathcal{E}_{dep} \text{ or }\\& (w_j, w_i) \in \mathcal{E}_{dep} \\
1 & \text{if } i = j \\
0 & \text{otherwise}
\end{cases}
\end{equation}

Here, $(w_i, w_j) \in \mathcal{E}_{dep}$ indicates a syntactic dependency between tokens $w_i$ and $w_j$. The matrix ensures bidirectional connectivity along dependency edges and self-loops at diagonal positions to retain token-specific features during graph convolution.

\textbf{Constituency Adjacency Matrix}: We adopt a bottom-up constituency parsing approach to capture phrase-level syntactic relationships. Constituency parsing organizes tokens into nested phrasal units (e.g., S, NP, VP), forming hierarchical graph connections that reflect both local and global syntactic patterns.

Given a constituency parse tree, we define depth-based layers $\mathcal{L} = \{L_d \mid d \in [0, D]\}$, where $L_d$ includes all constituent nodes at depth $d$ from the leaves, and $D$ is the tree's maximum depth. The multi-layer adjacency tensor $\mathbf{A}^{con} \in \{0,1\}^{l_c \times n \times n}$ is constructed as:

\begin{equation}
\mathbf{A}_{ij}^{con(d)} = \begin{cases}
1, & \text{if tokens } w_i, w_j \text{ belong to} \\
& \text{same constituent at depth } d \\
0, & \text{otherwise}
\end{cases}
\end{equation}

Here, $d$ is the depth level, and $l_c$ is the number of selected layers. We apply selective sampling at regular intervals to manage computational complexity, capturing both fine- and coarse-grained constituency relationships.

\textbf{Semantic Adjacency Matrix}: To capture contextual relationships between tokens, we construct a semantic adjacency matrix ${A}^{\text{sem}} \in \mathbb{R}^{n \times n}$ using multi-head self-attention, defined as:
\begin{equation}
{A}^{\text{sem}}_{ij} = \text{softmax} \left( \text{MHA}({h}_i^{bert}, {h}_j^{bert}) \right)
\end{equation}
where $\text{MHA}$ represents the multi-head attention score between tokens $i$ and $j$, and the softmax function normalizes the attention weights. Self-loops are added by assigning non-zero values to diagonal entries, and optional sparsification retains the most informative semantic links.

\subsection{Unified Graph Convolutional Networks}
Our framework employs a unified Graph Convolutional Network (GCN) to process four graph representations: constituency (ConGCN), dependency (DepGCN), semantic (SemGCN), and Abstract Meaning Representation (AMR-GCN). This unified approach ensures consistent feature propagation across different structural views while preserving representation-specific parameters.

\subsubsection{Common GCN Formulation}
For each graph structure, $X \in \{\text{con}, \text{dep}, \text{sem}, \text{amr}\}$, we apply the same GCN operation. Let $H^X = \{h_1^X, h_2^X, \ldots, h_n^X \}$ represent the node representations, where $h_i^X \in \mathbb{R}^d$ is the feature vector for token $i$. The $l$-th GCN layer updates each node as:
\begin{equation}
\begin{split}
h_i^{X,(l)} = \sigma\Bigg( \frac{1}{d_i^X + 1} \Big( \sum_j A_{ij}^X h_j^{X,(l-1)} W^{X,(l)} \\
\quad + h_i^{X,(l-1)} W^{X,(l)} \Big) \Bigg)
\end{split}
\end{equation}

where $W^{X,(l)}$ is the learnable weight matrix, $d_i^X = \sum_j A_{ij}^X$ is the node degree, and $\sigma(\cdot)$ denotes ReLU activation. The summation aggregates neighboring features while the second term preserves self-node information.
\subsubsection{Graph-Specific Processing and Knowledge Integration}

Each graph structure is processed independently using a unified GCN with view-specific layer depths. Specifically, the constituency, dependency, semantic, and AMR representations employ $l_c$, $l_d$, $l_s$, and $l_a$ layers, respectively, to capture hierarchical parses, multi-hop syntactic relations, contextual associations, and semantic roles. This design allows each view to model its structural characteristics while sharing a consistent encoding operation. External knowledge is incorporated as ${H}^{kg} = [{h}_1^{kg}, {h}_2^{kg}, \dots, {h}_n^{kg}]$.

\subsection{Hierarchical Cross-Modal Fusion Network}

After processing each graph structure via the unified GCN, a Hierarchical Fusion (HF) network integrates multi-view representations at three levels: local syntactic fusion, intermediate semantic integration, and global knowledge incorporation. This hierarchical approach preserves structural relationships while enabling effective cross-modal information exchange.
\subsubsection{Level 1: Local Syntactic Fusion}

At the local level, we integrate the constituency and dependency graph representations to capture complementary syntactic signals. The outputs from the final GCN layers for constituency and dependency are denoted by $h_i^{con,(l_c)}$ and $h_i^{dep,(l_d)}$, respectively. A gating mechanism is used to balance these two views:
\begin{equation}
G_{syn}(i) = \sigma(W_{syn}[h_i^{con,(l_c)}; h_i^{dep,(l_d)}])
\end{equation}

The fused syntactic representation is computed as:
\begin{align}
\tilde{h}_i^{syn} &= G_{syn}(i) \odot h_i^{con,(l_c)} + \notag \\
&\quad (1 - G_{syn}(i)) \odot h_i^{dep,(l_d)}
\end{align}

We then refine this vector using multi-head self-attention, followed by a residual connection and layer normalization:
\begin{align}
h_i^{syn} &= \text{LayerNorm} \Big( \tilde{h}_i^{syn} + \notag \\
&\quad \text{MHA}(\tilde{h}_i^{syn}, \tilde{h}_i^{syn}, \tilde{h}_i^{syn}) \Big)
\end{align}

The final representation $h_i^{\text{syn}}$ encodes both fine-grained syntactic and contextual relationships, providing a rich input for higher-level semantic fusion.

\subsubsection{Level 2: Intermediate Semantic Integration}

At the intermediate level, we fuse deep semantic information by combining Abstract Meaning Representation (AMR) features and semantic role attention outputs. Let $h_i^{sem,(l_s)}$ and $h_i^{amr,(l_a)}$ denote the outputs from the final GCN layers of the semantic and AMR graphs, respectively. We use a learnable gate to combine them:
\begin{align}
\tilde{h}_i^{sem} &= G_{sem}(i) \odot h_i^{sem,(l_s)} + \notag \\
&\quad (1 - G_{sem}(i)) \odot h_i^{amr,(l_a)}
\end{align}
To enable interaction between syntactic and semantic perspectives, we apply multi-head cross-attention between the fused semantic representation and the local syntactic embedding $h_i^{syn}$:
\begin{align}
\tilde{h}_i^{inter} &= \text{LayerNorm} \Big( \tilde{h}_i^{sem} + \notag \\
&\quad \text{MHA}(\tilde{h}_i^{sem}, h_i^{syn}, h_i^{syn}) \Big)
\end{align}

Finally, the concatenation of the local syntactic embedding, refined semantic embedding, and original contextualized BERT token embedding $h_i$ is passed through a feedforward network:

\begin{equation}
h_i^{inter} = \text{FFN} ( [h_i^{syn}; \tilde{h}_i^{inter}; {h}_i]) 
\end{equation}

The resulting $h_i^{inter}$ serves as a unified intermediate representation, incorporating both syntactic and semantic information.

\subsubsection{Level 3: Global Knowledge Integration}

At the global level, we incorporate external knowledge to enhance semantic understanding. The intermediate token representation $h_i^{inter}$ is fused with corresponding knowledge graph embeddings $h_i^{kg}$ using a learnable gate:
\begin{align}
\tilde{h}_i^{kg} &= G_{kg}(i) \odot h_i^{inter} + 
 (1 - G_{kg}(i)) \odot h_i^{kg}
\end{align}

The gated representation is refined through multi-head cross-attention to enable global information exchange, with the intermediate features as keys and values. This is followed by residual connection and layer normalization to produce the final global representation:
\begin{align}
h_i^{global} &= \text{LayerNorm} \Big( \tilde{h}_i^{kg} + \notag \\
&\quad \text{MHA}(\tilde{h}_i^{kg}, h_i^{inter}, h_i^{inter}) \Big)
\end{align}

The output $h_i^{global}$ serves as a comprehensive representation, integrating syntactic, semantic, and knowledge-based features for downstream tasks.
\subsubsection{Cross-Modal Global Fusion}
To enable efficient cross-modal interaction, we project both BERT and global graph features to a lower-dimensional space and apply cross-modal attention to capture interactions between textual and structural representations. The final hierarchical fusion combines all levels with learnable importance weights:
\begin{equation}
h_i^{final} = \alpha_1 h_i^{syn} + \alpha_2 h_i^{inter} + \alpha_3 h_i^{enhanced}
\end{equation}

where $\alpha_i = \text{softmax}(\boldsymbol{\alpha})_i$ are learnable hierarchical weights that adaptively balance contributions from different fusion levels, and $h_i^{enhanced}$ incorporates cross-modal interactions and feature enhancement through residual connections.

This hierarchical fusion architecture ensures that information flows progressively from fine-grained syntactic patterns to high-level semantic understanding, while preserving the structural inductive biases learned by each specialized GCN component. Appendix C provides further theoretical motivation for hierarchical fusion compared to flat fusion.

\subsection{Multi-view structure-aware Contrastive Learning}

Our framework incorporates three contrastive objectives to enhance complementary structural representations: syntactic-semantic alignment, AMR consistency, and knowledge graph alignment.

To ensure computational efficiency, we focus contrastive learning on semantically salient nodes rather than all token pairs. We compute an attention-based importance score for each token using the semantic attention matrix $\mathbf{A}^{sem} \in \mathbb{R}^{n \times n}$:
\begin{equation}
\text{importance}(i) = \frac{1}{n}\sum_{j=1}^{n} \mathbf{A}^{\text{sem}}_{ij} + \max_{j=1}^{n} \mathbf{A}^{\text{sem}}_{ij}
\end{equation}

This formulation captures both global connectivity (mean attention) and peak semantic relevance (max attention), identifying tokens that serve as effective anchors for structural learning. Following the multi-head pooling strategy in MP-GCN \cite{9691319}, we select the top-$k$ tokens per sentence where $k = \max(1, \lfloor (\log_{10}(\max(2, n)))^2 \rfloor)$. This sparse sampling approach is theoretically grounded in Bourgain's Theorem \cite{pmlr-v97-you19b}, which demonstrates that $\mathcal{O}(\log^2 n)$ landmark points suffice to preserve essential structural properties in metric spaces.

For an anchor token $i$, we define a unified contrastive loss that applies to both syntactic and AMR views. The loss encourages proximity to structurally relevant tokens while pushing unrelated ones apart, using a margin-based objective:
\begin{equation}
\mathcal{L}_{T}(i) = \frac{1}{\delta} \cdot \text{ReLU}\left(\bar{d}_{\text{pos}} - \bar{d}_{\text{neg}} + \gamma\right)
\end{equation}
where $T \in \{\text{syn}, \text{amr}\}$, $\bar{d}_{\text{pos}}$ and $\bar{d}_{\text{neg}}$ are the average distances to positive and negative samples, respectively, $\gamma$ is the margin, and $\delta$ is a normalization factor.

For the syntactic case ($T = \text{syn}$), positive samples include tokens connected via dependency edges, second-order constituency edges (present in the constituency graph but not in the dependency), and the anchor token itself. In the AMR case ($T = \text{amr}$), positive samples are tokens semantically connected to the anchor through AMR relations (i.e., where the AMR adjacency matrix value is non-zero), excluding the anchor token.

The knowledge graph (KG) contrastive loss is defined as:
\begin{equation}
\mathcal{L}_{\text{kg}} = \frac{1}{N} \sum_{i=1}^{N} \mathcal{L}_{\text{kg}}(h_i)
\end{equation}
\begin{equation}
\mathcal{L}_{\text{kg}}(h_i) = - \log \frac{\text{sim}(\text{proj}(\tilde{h}_i^{\text{text}}), h_i^{\text{kg}})}{\sum_{j=1}^{N} \text{sim}(\text{proj}(\tilde{h}_i^{\text{text}}), h_j^{\text{kg}})}
\end{equation}

Text features from the dependency GCN are projected into the KG space and L2-normalized before applying the InfoNCE loss. Tokens at the same position form positive pairs. Here, $\text{proj}(\cdot)$ represents the L2-normalized linear projection from text to KG embedding space, $\text{sim}(\cdot)$ is the cosine similarity, and $N$ is the total number of valid tokens across the batch.

Finally, the unified supervised contrastive loss combines all objectives with learnable balancing coefficients:
\begin{equation}
\mathcal{L}_{\text{scl}} = \lambda_{\text{syn}} \sum_{i \in \mathcal{I}} \mathcal{L}_{\text{syn}}(i) + \lambda_{\text{amr}} \sum_{i \in \mathcal{I}} \mathcal{L}_{\text{amr}}(i) + \lambda_{\text{kg}} \mathcal{L}_{\text{kg}}
\end{equation}
where $\mathcal{I}$ denotes the set of selected important nodes across all samples in the batch, and $\lambda_{\text{syn}}$, $\lambda_{\text{amr}}$, and $\lambda_{\text{kg}}$ are hyperparameters controlling the relative importance of syntactic, AMR, and knowledge graph alignment objectives.

\subsection{Training Objective}
For the primary aspect-based sentiment analysis (ABSA) task, we apply a standard cross-entropy loss over the final hierarchically fused representations. Given the final token representations $h_i^{final}$ from the hierarchical fusion network, we compute aspect-specific sentiment predictions through a classification head:
\begin{equation}
\mathcal{L}_{\text{CE}} = -\frac{1}{M} \sum_{m=1}^{M} \sum_{c=1}^{C} y_m^c \log p_m^c
\end{equation}
where $M$ is the number of training samples, $C$ is the number of sentiment classes, $y_m^c$ is the ground truth label (1 if sample $m$ belongs to class $c$, 0 otherwise), and $p_m^c$ is the predicted probability for class $c$ obtained through softmax normalization over the classification logits.

The complete training objective integrates both the task-specific cross-entropy loss and the multi-view contrastive regularization:
\begin{equation}
\mathcal{L}_{\text{total}} = \mathcal{L}_{\text{CE}} + \mathcal{L}_{\text{scl}}
\end{equation}

This unified objective enables the model to simultaneously learn discriminative representations for sentiment classification while enforcing structural consistency across multiple graph views. The contrastive component acts as a regularizer, encouraging the model to preserve complementary structural relationships encoded in syntactic dependencies, semantic roles, and external knowledge, ultimately leading to more robust and interpretable aspect-based sentiment analysis.

\section{Experiments}
In this section, we outline the experimental setup, including the datasets, implementation details, and baseline models used for comparison. We then present performance results under both basic and advanced evaluation settings. Finally, we provide a qualitative analysis by examining representative examples to gain deeper insights into the model's behavior and effectiveness.
\begin{table}[ht]
\centering
\scriptsize 
\setlength{\tabcolsep}{6pt} 
\begin{tabular}{l|cc|cc|cc}
\toprule
\textbf{Dataset} & \multicolumn{2}{c|}{\textbf{Positive}} & \multicolumn{2}{c|}{\textbf{Negative}} & \multicolumn{2}{c}{\textbf{Neutral}} \\
\cmidrule{2-7}
& Train & Test & Train & Test & Train & Test \\
\midrule
Restaurant14 & 2164 & 727 & 807 & 196 & 637 & 196 \\
Laptop14   & 976 & 337 & 851 & 128 & 455 & 167 \\
Twitter    & 1507 & 172 & 1528 & 169 & 3016 & 336 \\
\bottomrule
\end{tabular}
\caption{Sentiment class distribution for SemEval14, and Twitter datasets.}
\label{tab:dataset_stats}
\end{table}

\subsection{Datasets and Experimental Setup}
We experiment on three aspect-level sentiment datasets: Restaurant14, Laptop14, and Twitter from the SemEval 2014 ABSA challenge, removing conflicting sentiment labels. Dataset statistics are shown in Table~\ref{tab:dataset_stats}.

For AMR preprocessing, we use AMRBART \cite{bai-etal-2022-graph} for semantic parsing and \cite{blodgett-schneider-2021-probabilistic} for token alignment, with SuPar \cite{Zhang2020SuPar} for dependency and constituent tree parsing. 

To incorporate external knowledge, we use WordNet\cite{wordnet} as a lexical-semantic knowledge base, which is converted into relational triples encoding synonymy, antonymy, and hypernym–hyponym relations. Knowledge graph embeddings are trained on these triples using OpenKE\cite{openke} with DistMult, and the resulting embedding matrix serves as the external knowledge representation $h_{kg}$. Each input token is linked to its corresponding WordNet concept via lemma matching, and the associated embedding is retrieved from the trained model.

We use the bert-base-uncased model \citet{devlin-etal-2019-bert} with 768-dimensional hidden vectors and a sequence length of 100. For the three datasets (in the order of Table~\ref{tab:main_results}), the AMRGCN, DepGCN, ConGCN, and SemGCN layers are set to (1, 8, 6), (5, 2, 6), (3, 6, 3), and (7, 8, 6). Hyperparameters $\delta$ and $\gamma$ are set to 10 and 0.2. We train for 15 epochs, evaluating after each epoch. All experiments run on A100 GPUs, with each dataset taking an average of ~15 minutes. Performance is evaluated using Accuracy and macro-F1 scores.

\subsection{Main Results}
\begin{table*}[ht]
\centering
\small
\begin{tabular}{lcccccc}
\toprule
\textbf{Model} & \multicolumn{2}{c}{\textbf{Restaurant14}} & \multicolumn{2}{c}{\textbf{Laptop14}} & \multicolumn{2}{c}{\textbf{Twitter}}\\
\cmidrule{2-7}
                                            & Accuracy & Macro-F1 & Accuracy & Macro-F1 & Accuracy & Macro-F1\\
\midrule
BERT \cite{devlin-etal-2019-bert}            & 85.62 & 78.28 & 77.58 & 72.38 & 75.28 & 74.11 \\
dotGCN \cite{chen2022discrete}               & 86.16 & 80.49 & 81.03 & 78.10 & 78.11 & 77.00 \\
R-GAT \cite{wang-etal-2020-relational}       & 86.60 & 81.35 & 78.21 & 74.07 & 76.15 & 74.88 \\
KE-IGCN \cite{wan2023knowledge}              & 86.70 & 81.05 & 81.06 & 77.89 & - & -\\
Dual GCN \cite{li-etal-2021-dual-graph}      & 87.13 & 81.16 & 81.80 & 78.10 & 77.40 & 76.02 \\
SSEGCN \cite{zhang2022ssegcn}                & 87.31 & 81.09 & 81.01 & 77.96 & 77.40 & 76.02 \\
BiSyn-GAT \cite{liang2022bisyn}              & 87.49 & 81.63 & 82.44 & 79.15 & 77.99 & 76.80 \\
TextGT \cite{yin2024textgt}                  & 87.31 & 82.27 & 81.33 & 78.71 & 77.70 & 76.45 \\
S2GSL \cite{chen2024s}                       & 87.31 & \textbf{82.84} & 82.46 & 79.07 & 77.84 & \textbf{77.11} \\
Bert+CD \cite{tian2024aspect}                & 87.32 & 81.94 & 82.25 & 79.65 & - & - \\
DMAN \cite{chen2024dynamic}                  & 87.59 & 82.47 & 82.29 & 78.91 & - & - \\
MambaforGCN+BERT \cite{lawan2025enhancing}   & 86.68 & 80.86 & 81.80 & 78.59 & 77.67 & 76.88 \\
\textbf{CMV-Fuse (ours)}                    & \textbf{87.76} & 81.99 & \textbf{82.71} & \textbf{79.79} & \textbf{78.13} & 77.06 \\
\bottomrule
\end{tabular}
\caption{Performance comparison of CMV-Fuse and baselines on the Restaurant14, Laptop14, and Twitter datasets using Accuracy and Macro-F1}
\label{tab:main_results}
\end{table*}

Table~\ref{tab:main_results} shows the experimental results on the SemEval-2014 datasets. CMV-Fuse demonstrates competitive performance across all three datasets. In Restaurant14, Laptop14 and Twitter, it beats all the baseline models in accuracy (87.76\%, 82.71\%, 78.13\%) and achieves comparable Macro-F1 (81.99\%, 79.79 77.06\%). This highlights the effectiveness of our Cross Modal-View Fusion in capturing aspect-sentiment relationships.


Compared to syntax-centric graph models such as Dual GCN (87.13\%) and dotGCN (86.16\%), CMV-Fuse improves accuracy on Restaurant14 by 0.63 percentage points and 1.60 percentage points, respectively, demonstrating the benefit of incorporating semantic abstraction and external knowledge beyond purely syntactic structures. Compared with BiSyn-GAT, which jointly models constituency and dependency trees, CMV-Fuse achieves consistent gains of 0.27 percentage points on accuracy for Restaurant14 and 0.14 percentage points for Twitter, highlighting the advantage of systematic multi-view integration over syntax-only dual-graph designs. CMV-Fuse also remains competitive with recent semantic and graph enhanced SOTA models, outperforming TextGT by 0.45 percentage points on accuracy for Restaurant14 and S2GSL by 0.40 percentage points on Twitter, while achieving the best overall performance on Laptop14 (82.71\% accuracy, 79.79\% Macro-F1). Overall, these results indicate that integrating AMR semantics, syntactic structure, semantic attention, and external knowledge enables more robust aspect-level sentiment reasoning across diverse datasets.


We conduct comprehensive ablation studies (Appendix B) to isolate and understand the individual contribution of each view (AMR, DEP, CON, SEM, KG), effectiveness of complementary view pairings and impact of different training objectives (e.g., alignment and contrastive losses) on performance gains. This analysis clarifies which components drive performance gains and avoids attributing improvements to the framework without evidence (results in Appendix B Tables \ref{tab:fusion_results} and \ref{tab:loss_results}). Results show that combining all three views (AMR, syntax, and knowledge) achieves the best performance (87.76\% accuracy). Contrastive learning, especially for semantic consistency, is crucial for model performance.

\section{Conclusion}

This work presents CMV-Fuse, a unified modular ABSA framework that integrates multiple linguistic views, including AMR semantics, dependency and constituency syntax, semantic attention, and external knowledge. Unlike prior methods that rely on isolated structures, CMV-Fuse uses a hierarchical gated fusion mechanism to jointly model local syntactic cues, intermediate semantic representations, and global knowledge signals, aligning with the layered nature of language understanding. A multi-view contrastive learning objective further aligns complementary representations and improves cross-view consistency with low overhead. Experiments show that CMV-Fuse consistently outperforms strong baselines, while ablation studies confirm the complementary contributions of each linguistic view and the model’s robustness across diverse sentiment settings. Overall, CMV-Fuse advances more integrated, interpretable, and linguistically grounded sentiment analysis, with potential for cross-domain, multilingual, and fairness-aware extensions.

\section{Limitation}
While our model offers flexibility by integrating multiple linguistic representations (e.g., dependency, constituency, AMR, and KG-based views), it also presents certain limitations. First, the effectiveness of individual representations varies depending on dataset characteristics. For instance, datasets such as Restaurants, which contain shorter and more aspect-specific samples, may not benefit from complex semantic or long-range dependency modeling; a simple syntactic structure (e.g., dependency parsing) is often sufficient. Conversely, datasets like Laptops typically involve longer and semantically richer sentences that require more advanced representations, such as AMR or contextualized embeddings. Consequently, determining which representations meaningfully contribute to a specific dataset remains a challenge.

Another limitation lies in the reliance on off-the-shelf parsers for generating linguistic views (e.g., dependency, constituency, or AMR parses). While these tools simplify preprocessing, their inherent parsing errors or domain mismatches can propagate through the model and affect overall performance. Additionally, the KG-based component depends on the availability and quality of an external knowledge graph tailored to the target domain, which may not always exist or be easy to construct.

Finally, the current evaluation is limited to English benchmark datasets; future work should explore multilingual and domain-general extensions, as well as deeper interpretability analyses to better understand how each view contributes to the final prediction.

\bibliography{custom}

\appendix
\section{Appendix A: Baseline Methods}
We compare CMV-Fuse with several state-of-the-art models, including: (1) \textbf{BERT} \cite{devlin-etal-2019-bert}, a general pre-trained model with a classification head for ABSA; (2) \textbf{R-GAT} \cite{wang-etal-2020-relational}, which uses a relational GAT for aspect-based dependency structures; (3) \textbf{DualGCN} \cite{li-etal-2021-dual-graph}, a dual GCN that merges features from sentences and dependency trees; (4) \textbf{dotGCN} \cite{chen2022discrete}, which uses a latent tree structure for aspects; and (5) \textbf{SSEGCN} \cite{devlin-etal-2019-bert}, incorporates aspect-aware attention in GCNs.(6) \textbf{KE-IGCN} \cite{wan2023knowledge} aims to select the highly relevant subgraphs, and proposes an interaction strategy to evaluate the interaction between external knowledge and the input text. (7) \textbf{BiSyn-GAT} \cite{liang2022bisyn} graph-based model that captures syntactic dependencies in both bottom-up and top-down directions using attention mechanisms, enabling richer contextual representations for each token. (8) \textbf{TextGT} employs a double-view graph transformer to integrate syntactic and semantic dependencies for ABSA. (9) \textbf{S2GSL} models syntactic graphs with graph structure learning to improve aspect–opinion interaction. (10) \textbf{Bert+CD} enhances BERT with contextualized dependency(CD) representation, injecting syntactic dependency information into token embeddings. (11) \textbf{DMAN}, a Dual Mode Attention Network that jointly models semantic and syntactic information through complementary attention mechanisms. (12) \textbf{MambaforGCN+Bert} integrates the Mamba sequencing modeling architecture with GCN and BERT, leveraging long range dependency modeling from Mamba and structured syntactic reasoning from GCN.

\section{Appendix B: Ablation study}
\subsection{Ablation Study}

We conduct comprehensive ablation studies to isolate and understand
\begin{enumerate}
    \item the individual contribution of each view (AMR, DEP, CON, SEM, KG)
    \item the effectiveness of complementary view pairings
    \item the impact of different training objectives (e.g., alignment and contrastive losses) on performance gains.
    \item The baselines considered included SOTA from mainly the last 3 years (Table \ref{tab:main_results})
\end{enumerate}

This analysis clarifies which components drive performance gains and avoids attributing improvements to the framework without evidence (results in Appendix Table \ref{tab:fusion_results} and Table \ref{tab:loss_results}).


We conduct comprehensive ablation studies to validate the contribution of each component in CMV-Fuse, examining multi-view combinations, contrastive learning objectives, and architectural design choices. All experiments are conducted on Restaurant14 unless otherwise specified.

\begin{table}[ht]
\centering
\setlength{\tabcolsep}{5pt}
\begin{tabular}{ccc|cccc}
\toprule
$H_{amr}$ &  $H_{syn}$ & $H_{kg}$ & Acc. (\%) & F1 (\%) \\
\midrule
\checkmark &  &                & 86.91 & 79.94  \\
        & \checkmark &         & 86.26 & 79.65  \\
        &         & \checkmark & 86.54 & 79.69  \\
\checkmark & \checkmark &         & 87.01 & 80.73 &  \\
\checkmark &         & \checkmark & 87.10 & \textbf{80.87} \\
        & \checkmark & \checkmark & 86.54 & 80.62  \\
\checkmark & \checkmark & \checkmark & \textbf{87.23} & 80.82 \\
\bottomrule
\end{tabular}
\caption{Ablation study of Different Multi View Fusion Combinations on the Restaurant14 dataset.}
\label{tab:fusion_results}
\end{table}

\subsubsection{Effects of Different Multi View Fusion Combinations}
Table~\ref{tab:fusion_results} systematically evaluates all combinations of our three core representations: AMR-based semantic ($H_{\text{amr}}$), syntactic structures ($H_{\text{syn}}$), and external knowledge ($H_{\text{kg}}$). Individual views show that $H_{\text{amr}}$ achieves the highest standalone performance (86.91\% accuracy), validating semantic abstraction's discriminative power for ABSA. Among pairwise combinations, $H_{\text{amr}} + H_{\text{kg}}$ (87.10\%) performs best, demonstrating effective knowledge-semantic synergy, while $H_{\text{syn}} + H_{\text{kg}}$ (86.54\%) shows minimal gains, indicating syntactic structures require semantic abstraction to effectively leverage external knowledge. 

For fair comparison, we concatenate different representations and feed them to a single-layer MLP classifier to achieve multi-view fusion. Note that we do not employ neither the proposed hierarchical fusion module nor any contrastive loss objectives to fuse the entire representation combination [$H_{amr}$,$H_{syn}$,$H_{kg}$]; thus, the performances of the last row in Table \ref{tab:fusion_results} are slightly worse than the relative ones in Table \ref{tab:main_results}. The full three-way combination achieves optimal results (87.23\%), confirming that systematic multi-view integration provides the most robust representation. 

\begin{table}[ht]
\centering
\setlength{\tabcolsep}{5pt}
\begin{tabular}{c|cccc}
\toprule
Model & Acc. (\%) & F1 (\%) \\
\midrule
Our CMV-Fuse        & \textbf{87.76} & \textbf{81.99}  \\
\midrule
\textit{W/O} $\mathcal{L}_{scl}$ & 87.38 & 81.72  \\
\textit{W/O} $\mathcal{L}_{amr}$ & 86.82 & 79.11  \\
\textit{W/O} $\mathcal{L}_{syn}$ & 86.54 & 80.00 &  \\
\textit{W/O} $\mathcal{L}_{kg}$ & 87.01 & 80.62 \\

\bottomrule
\end{tabular}
\caption{Ablation study of Different Multi View Loss Contributions on the Restaurant14 dataset}
\label{tab:loss_results}
\end{table}

\subsubsection{Effects of Different Multi View Loss Contributions}
Table~\ref{tab:loss_results} reveals that contrastive learning is essential: removing $\mathcal{L}_{\text{amr}}$ causes the largest degradation (-0.94 percentage points for accuracy and -2.88 percentage points for F1), demonstrating critical importance of semantic consistency, while eliminating all contrastive objectives results in severe performance loss (-0.38 percentage points for accuracy, -0.27 percentage points for F1). Our full model with hierarchical fusion and all contrastive objectives achieves 87.76\% accuracy and 81.99\% F1, significantly outperforming the concatenation baseline of 87.23\% (+0.53 percentage points) accuracy and 80.82\% (+1.17 percentage points) F1.

\end{document}